\documentclass[conference]{IEEEtran}
\usepackage{amsmath,graphicx}
\usepackage{url}
\usepackage{xcolor}
\definecolor{SoftWildStrawberry}{RGB}{255, 120, 180} 
\usepackage[
    colorlinks=true,
    linkcolor=black,
    citecolor=black,
    urlcolor=black
]{hyperref}
\usepackage[numbers]{natbib}  

\title{Investigating the Effect of Network Pruning on Performance and Interpretability}
\author{\IEEEauthorblockN{Jonathan von Rad, Florian Seuffert}
\IEEEauthorblockA{AI Center, Neural Information Processing Group\\
University of Tübingen, Tübingen, Germany}
}

\begin{document}

\maketitle

\begin{abstract}
Deep Neural Networks (DNNs) are often over-parameterized for their tasks and can be compressed quite drastically by removing weights, a process called \emph{pruning}. We investigate the impact of different pruning techniques on the classification performance and interpretability of GoogLeNet. We systematically apply unstructured and structured pruning, as well as connection sparsity (pruning of input weights) methods to the network and analyze the outcomes regarding the network's performance on the validation set of ImageNet. We also compare different retraining strategies, such as iterative pruning and one-shot pruning. We find that with sufficient retraining epochs, the performance of the networks can approximate the performance of the default GoogLeNet---and even surpass it in some cases. To assess interpretability, we employ the Mechanistic Interpretability Score (MIS) developed by Zimmermann et al.\ \cite{zimmermann2024measuring}. Our experiments reveal that there is no significant relationship between interpretability and pruning rate when using MIS as a measure. Additionally, we observe that networks with extremely low accuracy can still achieve high MIS scores, suggesting that the MIS may not always align with intuitive notions of interpretability, such as understanding the basis of correct decisions. 
Code is available at \href{https://github.com/Uni-Tubingen-Teamproject/Network-Pruning-and-Interpretability}{\texttt{\color{SoftWildStrawberry}{Github}}}.
\end{abstract}

\IEEEpeerreviewmaketitle

\section{Introduction}

\noindent
The rapid advancement of deep learning over the past decade has led to the development of increasingly complex neural network architectures, such as GoogLeNet (often called Inception V1)~\cite{szegedy2015going}, ResNet~\cite{he2016deep}, and others. While these models have achieved remarkable performance across various tasks, their large size and computational demands present significant challenges, particularly when deploying them in resource-constrained environments. Neural network pruning has emerged as a vital technique---inspired by the foundational work of LeCun et al.~\cite{lecun1990optimal}---to address these challenges by reducing model size while attempting to retain as much accuracy as possible.

Pruning methods have been explored extensively in the literature. Early work by Han et al.~\cite{han2015learning} introduced the concept of unstructured pruning, where individual weights are pruned based on their magnitude, leading to significant reductions in model sizes. For example, in their work with AlexNet, they achieved a 9x reduction in the number of parameters with only a negligible drop in Top-1 accuracy from 57.2\% to 57.1\%. However, unstructured pruning does not guarantee speedups in practice due to the irregular sparsity patterns it creates, which are difficult to exploit efficiently on standard hardware.

More recent research has focused on structured pruning, which involves pruning entire filters, channels, or layers, resulting in more regular sparsity patterns that are easier to accelerate on modern hardware. For instance, Li et al.~\cite{li2017pruning} demonstrated that pruning 34\% of the filters in VGG-16 led to a reduction in the number of parameters with less than a 0.1\% drop in Top-5 accuracy on ImageNet. Similarly, Molchanov et al.~\cite{molchanov2017pruning} showed that by pruning 64\% of the parameters in VGG-16, they achieved a minimal loss of 0.6\% in Top-1 accuracy. These results underscore the potential of structured pruning to produce compact models with minimal accuracy degradation.

Building upon these approaches, we applied them to GoogLeNet and explored an alternative pruning strategy we termed Connection Sparsity. This method focuses on pruning input channels, which differs from traditional structured pruning that targets output channels. By pruning input channels, Connection Sparsity preserves the structural integrity of the network while still reducing the number of connections, which is beneficial for optimizing inference speeds. Previous studies, such as those by Li et al.~\cite{li2017pruning} and He et al.~\cite{he2017channel}, have suggested that such approaches can lead to more efficient models with less accuracy degradation. Remarkably, our experiments demonstrated that applying Connection Sparsity to GoogLeNet allowed us to prune 80\% of the entire model while achieving an increase in Top-1 accuracy on ImageNet by 0.2\% after retraining. To the best of our knowledge, this result is notable among the pruning settings considered here, highlighting the effectiveness of this approach.

Another significant aspect of our research is the comparison between iterative and one-shot pruning techniques. Iterative pruning has been shown to be more effective in maintaining model accuracy, as it allows the network to gradually adapt to the reduced number of parameters~\cite{zhu2017prune, frankle2018lottery}. However, it is more computationally expensive compared to one-shot pruning, where a large portion of the network is pruned in a single step and retrained afterwards. Our work evaluates the trade-offs between these two approaches, showing that while iterative pruning yields slightly better results, it comes at a significant computational cost.

While much of the existing literature has focused on the impact of pruning on performance metrics like accuracy and space complexity, the effect of pruning on interpretability has received less attention. Interpretability is crucial for understanding and trusting model decisions, especially in critical applications. Few recent works have started to explore the interpretability of neural networks post-pruning~\cite{hooker2021compressed, frankle2019dissecting}. The Mechanistic Interpretability Score (MIS) introduced by Zimmermann et al.~\cite{zimmermann2024measuring} provides a quantifiable measure of interpretability based on the perceptual similarity between explanations and queries, both coming from the dataset the network is trained on. However, as our findings suggest, the MIS indicates that pruned networks might become more interpretable, though there are methodological concerns, especially regarding the scores in the softmax layer, which may not accurately reflect true interpretability.

Our research builds on these previous studies by systematically examining the impact of different pruning techniques---Unstructured, Structured, and Connection Sparsity---on both validation accuracy performance and the MIS as a measure of interpretability in pruned networks. Moreover, we propose the need for more robust metrics that align better with decision-making accuracy, as existing measures may not fully capture the nuances of model interpretability post-pruning.

This study not only contributes to the ongoing efforts to develop more efficient and interpretable deep learning models but also highlights the limitations of existing interpretability metrics when applied to pruned networks. By addressing these gaps, our work provides a foundation for future research aimed at improving the deployment of neural networks in real-world applications where both performance and interpretability are critical.

\section{Effect on Performance}

\noindent
To measure the effect of pruning on performance, we pruned GoogLeNet using three different approaches: Unstructured, Structured, and Connection Sparsity pruning. Each pruning method was evaluated with both iterative and one-shot pruning scenarios.

\subsection{Pruning Strategies}

\begin{itemize}
    \item \textbf{Unstructured Pruning}: Removes weights across the entire network, leading to a sparse network without any specific structural pattern.
    \item \textbf{Structured Pruning}: Removes entire filters or channels, leading to a reduced but dense network.
    \item \textbf{Connection Sparsity}: Focuses on pruning input channels, preserving the structural integrity of the network while reducing the number of connections between those.
\end{itemize}

\subsection{Iterative vs. One-Shot Pruning}

\noindent
Our results align with the findings of Frankle and Carbin~\cite{frankle2018lottery}, who proposed in the Lottery Ticket Hypothesis that iterative pruning is more effective than one-shot pruning, particularly in maintaining performance after significant pruning. In our experiments with GoogLeNet, we observed similar outcomes, where iterative pruning yielded slightly better results. However, this advantage was only evident when the network was retrained to convergence after each pruning step, which in the case of GoogLeNet, required approximately 50 epochs per retraining. The necessity of this amount of epochs is evident in Figure~\ref{fig:accuracy_epochs}. While this approach can lead to marginally better performance, it is computationally inefficient.

\subsection{Results}

\noindent
Our results show that Connection Sparsity pruning outperformed both Unstructured and Structured pruning methods in preserving accuracy while significantly reducing the network size. Iterative pruning, when applied, consistently provided better accuracy compared to one-shot pruning.

\begin{figure}[!h]
    \centering
    \includegraphics[width=0.5\textwidth]{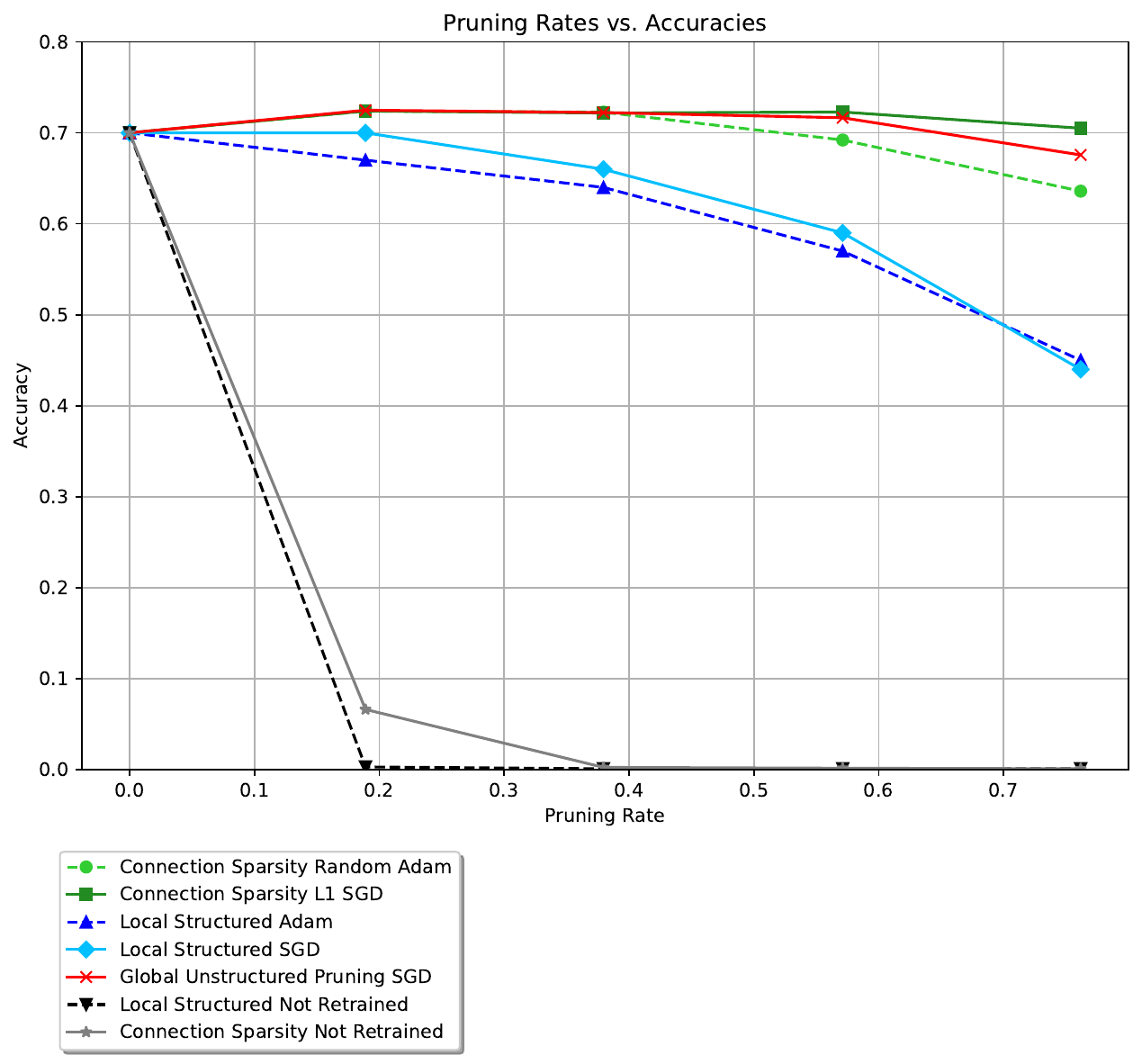}
    \caption{Summary of Pruning Methods and Their Impact on Accuracy. Black lines represent models without fine-tuning.}
    \label{fig:summary_pruning_methods}
\end{figure}

For each setup, we experimented with various hyperparameters, including the learning rate, scheduler, and the number of training epochs. After testing different configurations, we found the following settings to be the most effective: optimizer=SGD, scheduler=ExponentialLR, learning rate=0.01, and 50 training epochs. We also experimented with the Adam optimizer with a learning rate of 0.001, but this did not result in significant differences, as shown in Figure~\ref{fig:summary_pruning_methods}.

\begin{figure}[!h]
    \centering
    \includegraphics[width=0.5\textwidth]{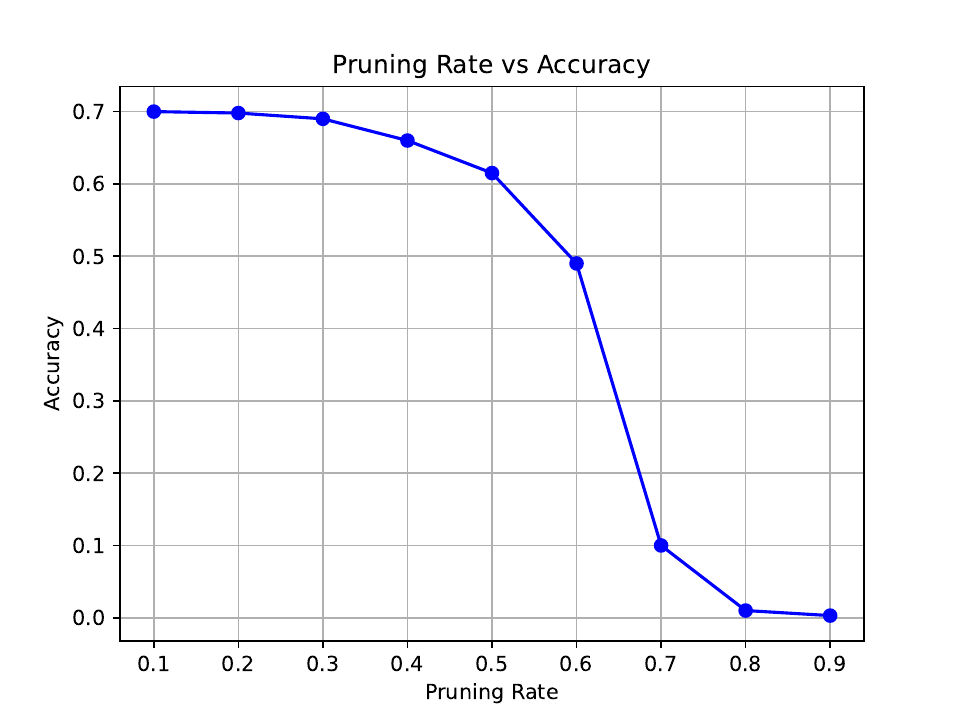}
    \caption{Pruning Rate vs. Accuracy for Global Unstructured Pruning (L1 norm) without fine-tuning.}
    \label{fig:pruning_vs_accuracy}
\end{figure}

Figure~\ref{fig:pruning_vs_accuracy} illustrates the global unstructured pruning with L1 norm without fine-tuning. As seen, the accuracy drastically drops with higher pruning rates, emphasizing the need for fine-tuning.

\begin{figure}[!h]
    \centering
    \includegraphics[width=0.5\textwidth]{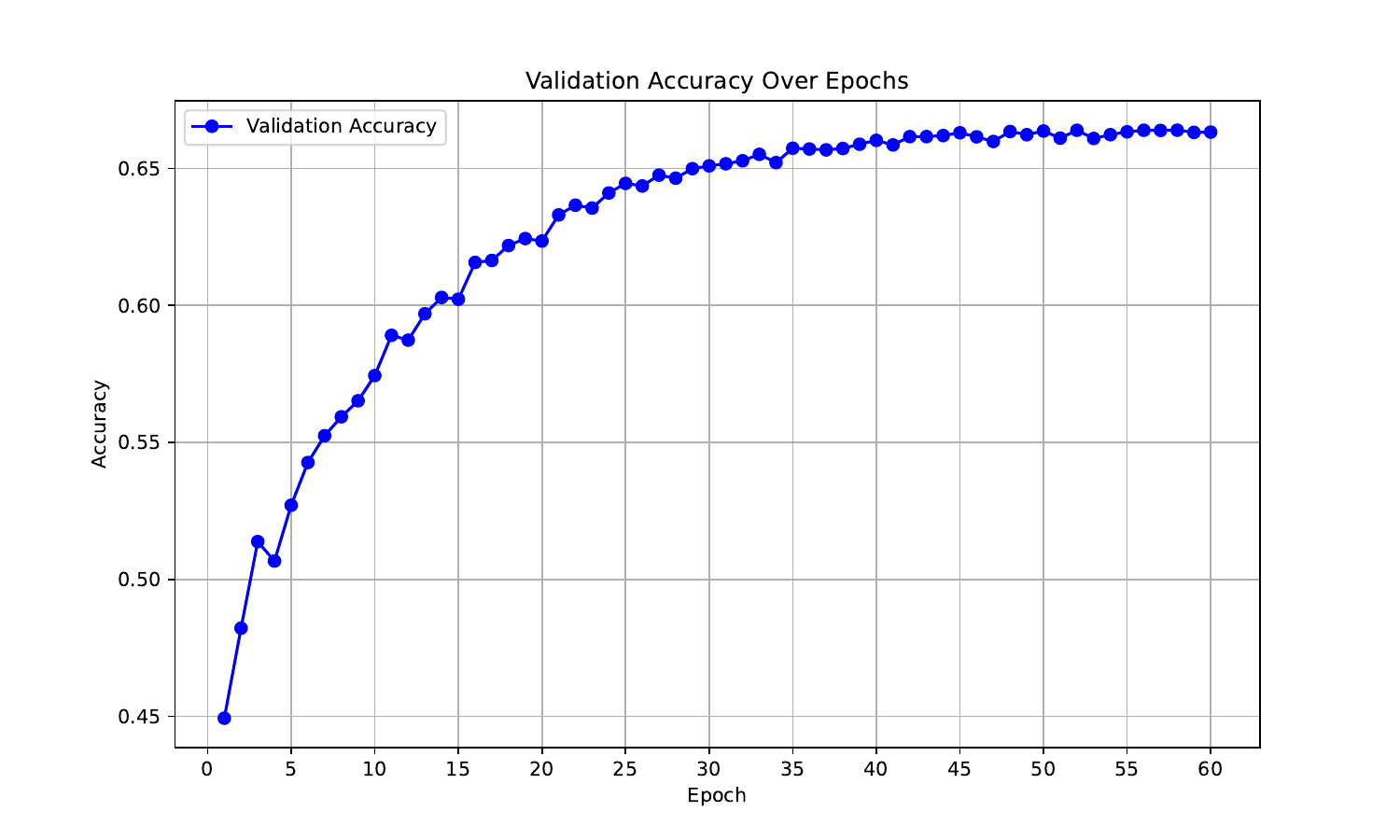}
    \caption{Validation Accuracy Over 60 Epochs After Unstructured Pruning 89\% of the Network (One-Shot Retraining).}
    \label{fig:accuracy_epochs}
\end{figure}

In Figure~\ref{fig:accuracy_epochs}, we present the development of accuracy over 60 epochs of one-shot retraining after unstructured pruning 89\% of the network. The network gradually regains accuracy as it adapts to the reduced parameters.

\begin{figure}[!h]
    \centering
    \includegraphics[width=0.5\textwidth]{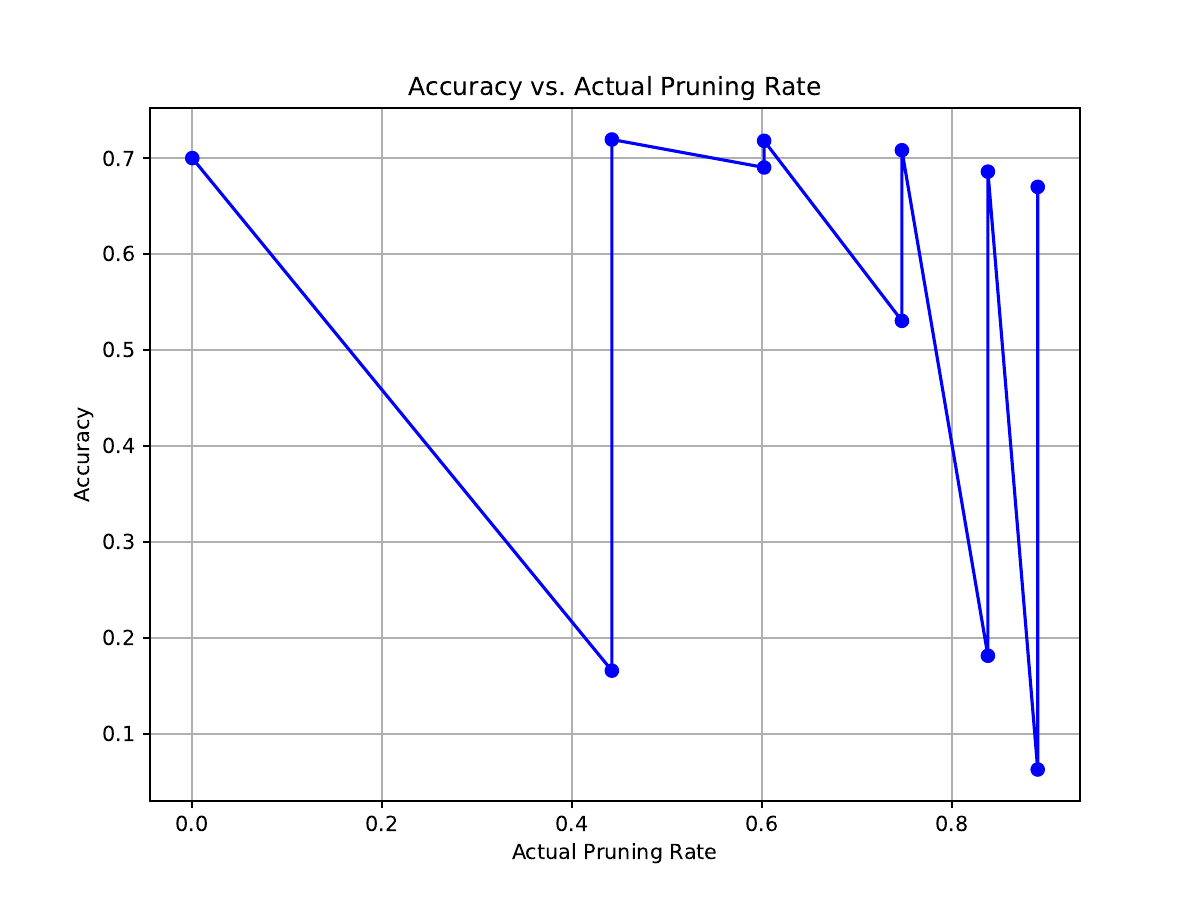}
    \caption{Accuracy vs. Pruning Rate: Iterative Pruning with Fine-Tuning vs. One-Shot Pruning.}
    \label{fig:iterative_vs_one_shot}
\end{figure}

Figure~\ref{fig:iterative_vs_one_shot} contrasts the accuracy results between one-shot pruning (Figure~\ref{fig:accuracy_epochs}) and iterative pruning and fine-tuning with the same pruning rate, showing how iterative pruning yields higher accuracy (68\% for iterative compared to 66\% for one-shot pruning).

Figure~\ref{fig:summary_pruning_methods} summarizes the performance of the three pruning methods: Unstructured (red), Structured (blue), and Connection Sparsity (green), after pruning and fine-tuning. Notably, Unstructured Pruning and the Connection Sparsity method showed even higher accuracy values than before pruning, while the Connection Sparsity approach, which involves pruning input channels, also offers optimization benefits due to its structured nature. Surprisingly, Connection Sparsity performs remarkably well even when combined with random pruning---a result that is not observed with either unstructured or structured (output channel) pruning methods.

Due to concerns that the performance improvements were only due to a high number of retraining epochs, we also trained a standard GoogLeNet to disprove this hypothesis. After 50 additional training epochs, the standard GoogLeNet showed only minor improvements and achieved a validation accuracy of 72.1\%. Still, this is below the highest accuracy achieved with the pruned and then retrained models, especially those with sparse connections where we achieved validation accuracies of almost 72.3\% (72.286\%) with 60\% of all parameters pruned. Even with 80\% of all parameters set to 0 via Connection Sparsity, the retrained network out-performed the default GoogLeNet regarding its performance on the validation set with an accuracy of 70.5\%.

\section{Effect on Interpretability}

\noindent
To assess the interpretability of pruned networks, we utilized the Mechanistic Interpretability Score (MIS) as proposed by Zimmermann et al.~\cite{zimmermann2024measuring}. This score is designed to measure how well an observer can understand the cause of a network's decision.

\subsection{Mechanistic Interpretability Score (MIS)}

\noindent
The MIS quantifies interpretability based on the perceptual similarity between explanations (natural images that maximize or minimize filter activations) and queries (images that elicit strong or weak activations from the unit). While this metric provides a way to evaluate interpretability, it does not necessarily capture all aspects of how understandable a network's decision-making process is.

\subsection{Considerations Regarding MIS}

\noindent
Our experiments demonstrated that there is no significant relationship between interpretability and pruning rate when using the MIS as a measure. Additionally, we observed that networks with extremely low accuracy (e.g., 0.1\%) could still achieve high MIS scores. This finding might initially seem counterintuitive, as a high level of interpretability is often associated with an enhanced understanding of correct decisions. However, this is not necessarily a contradiction under the definition of interpretability assumed by the MIS. For instance, a network that primarily distinguishes colors may have almost random overall performance but still be highly interpretable in terms of knowing which units will activate for a given input (i.e., high similarity between high-activating images). Thus, a network may be considered "interpretable" by the MIS metric even if it does not make meaningful overall predictions.

\begin{figure}[!h]
    \centering
    \includegraphics[width=0.48\textwidth]{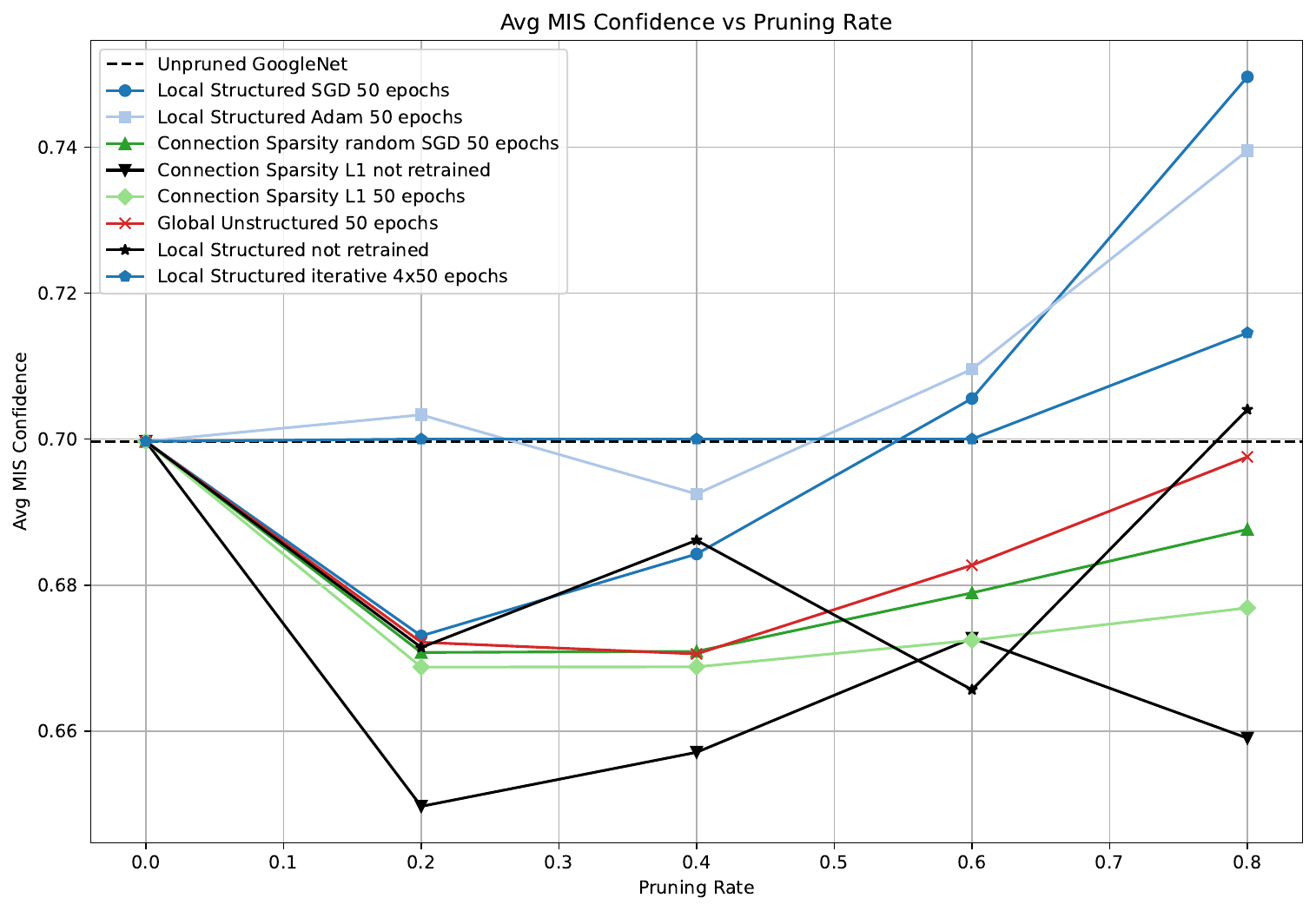}
    \caption{Average MIS Confidence vs. Pruning Rate. The black lines represent models without fine-tuning and therefore with near-random predictions.}
    \label{fig:mis_vs_pruning}
\end{figure}

Figure~\ref{fig:mis_vs_pruning} illustrates the relationship between the pruning rate and interpretability, as measured by the MIS metric. The correlation between interpretability and pruning rate is 0.092, indicating no significant relationship between the two. The high MIS scores observed for models making random predictions (represented by the black lines) suggest that the MIS may not always align with other notions of interpretability, such as understanding the basis of correct decisions.

An additional observation from our analysis is that the MIS score for the softmax layer was consistently low across all models. This result was unexpected given the methodology behind MIS, which suggests that the interpretability of class units in the softmax layer should be high. Since explanations and queries in this layer should predominantly involve class members, one would anticipate a high MIS score.

\begin{figure}[!h]
    \centering
    \includegraphics[width=0.5\textwidth, trim={0.5cm 0.5cm 0.5cm 0.5cm}, clip]{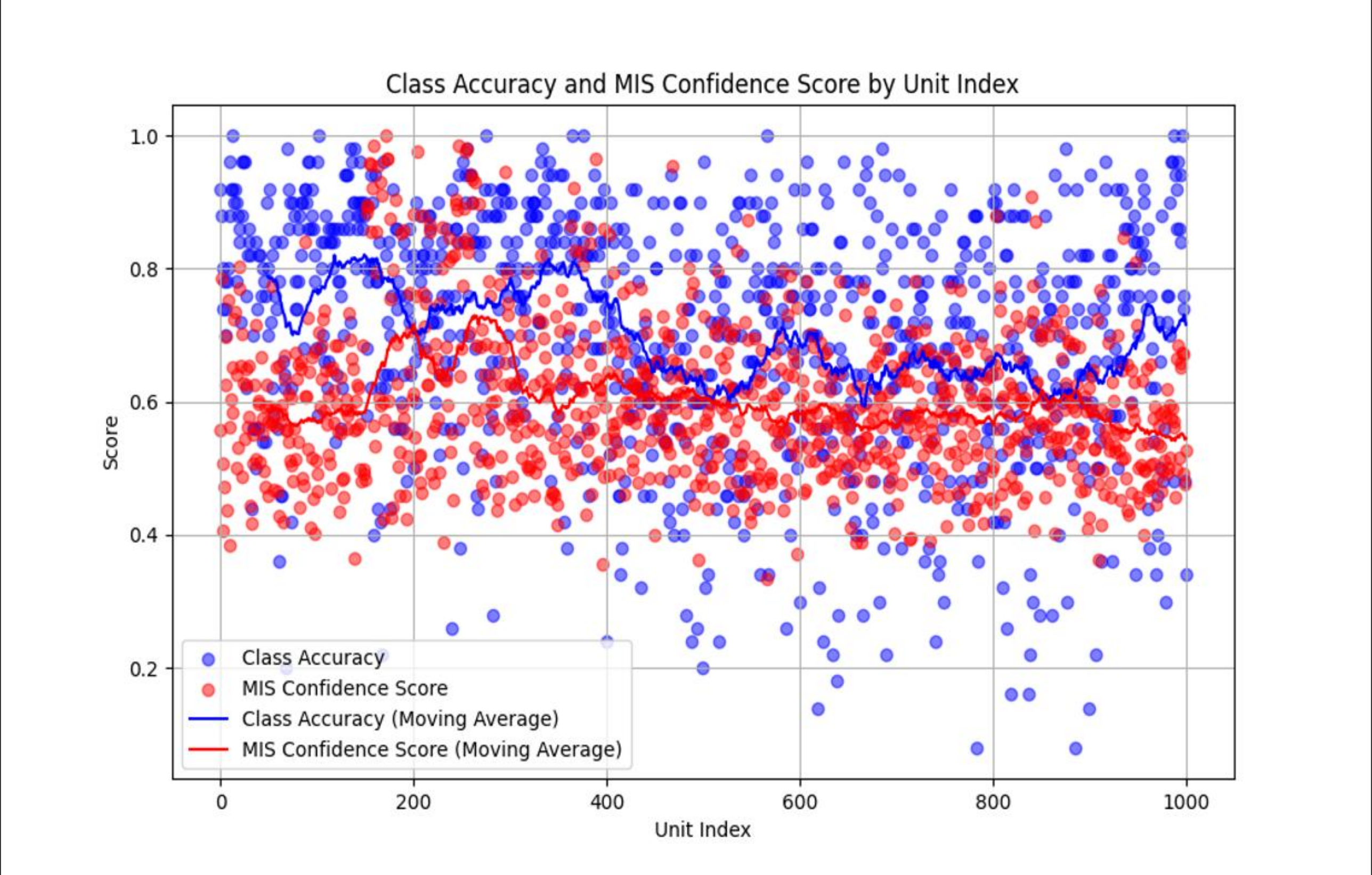}
    \caption{Class accuracy (blue) and MIS confidence score (red) for different units in a local structured 20\% pruned and 50 epochs retrained model. Solid lines indicate moving averages for each metric.}
    \label{fig:class_accuracy_mis}
\end{figure}

To investigate this further, we computed the correlation between the class-wise accuracy of each unit in the softmax layer and their respective MIS scores. If a class is well-recognized by its unit in the softmax layer, it is expected that explanations would mostly involve class members, and queries should be easier to assign, resulting in a high MIS score. However, for three pruned models (Connection Sparsity with 20\% of parameters pruned and 50 epochs of retraining; Structured Pruning with 20\% of parameters pruned and 50 epochs of retraining; Structured Pruning with 80\% of parameters pruned and 50 epochs of retraining), there was no significant correlation between class-wise accuracy and MIS scores (all correlation coefficients were below 0.09).

\section{Conclusion}

\noindent
In this study, we have shown that different pruning strategies can have varying effects on both the performance and interpretability of GoogLeNet. Connection Sparsity pruning, in combination with iterative retraining, offers the best balance between model compression and accuracy retention. However, the MIS, as a measure of interpretability, does not always provide meaningful insights, especially when applied to networks with low accuracy. This is not necessarily a flaw in the metric but reflects the nuances of what "interpretability" might entail. Future research should focus on developing more reliable interpretability metrics that consider both the clarity of the decision-making process and the quality of decisions made by the network.

\bibliographystyle{IEEEtran}
\bibliography{main}

\end{document}